\documentclass{article}
\usepackage{ijcai20}

\usepackage{times}
\usepackage{soul}
\usepackage{url}

\usepackage[utf8x]{inputenc}
\usepackage[small]{caption}
\usepackage{graphicx}
\usepackage{epstopdf}
\usepackage{amsmath}
\usepackage{booktabs}
\usepackage{algorithm}
\usepackage{algorithmic}
\newtheorem{theorem}{Theorem}
\usepackage{xcolor}
\usepackage{subfigure}

\title{FOCUS: Dealing with Label Quality Disparity in Federated Learning}

\author{\#8163}
\author{
Yiqiang Chen$^{1,2}$\footnote{Corresponding Author}\and
Xiaodong Yang$^{1,2}$\and
Xin Qin$^{1,2}$\and
Han Yu$^{3}$\and
Biao Chen$^{4}$ \and
Zhiqi Shen$^{3}$ 
\\
\affiliations
$^1$The Beijing Key Laboratory of Mobile Computing and Pervasive Device, Institute of Computing Technology, Chinese Academy of Sciences, Beijing, China\\
$^2$University of Chinese Academy of Sciences, Beijing, China\\
$^3$Nanyang Technological University, Singapore\\
$^4$Xuanwu Hospital, Capital Medical University, Beijing, China
\emails
\{yqchen, yangxiaodong, qinxin18b\}@ict.ac.cn, han.yu@ntu.edu.sg, pbchan@hotmail.com, zqshen@ntu.edu.sg
}

\begin{document}

\maketitle

\begin{abstract}
Ubiquitous systems with End-Edge-Cloud architecture are increasingly being used in healthcare applications. Federated Learning (FL) is highly useful for such applications, due to silo effect and privacy preserving. Existing FL approaches generally do not account for disparities in the quality of local data labels. However, the clients in ubiquitous systems tend to suffer from label noise due to annotators’ varying skill-levels, biases or malicious tampering. In this paper, we propose Federated Opportunistic Computing for Ubiquitous Systems (FOCUS) to address this challenge. It maintains a small set of benchmark samples on the FL server and quantifies the credibility of the clients’ local data without directly observing them by computing the mutual cross-entropy between performance of the FL model on the local datasets and that of the client’s local FL model on the benchmark dataset. Then, a credit-weighted orchestration is performed to adjust the weight assigned to clients in the FL model based on their credibility values. FOCUS has been experimentally evaluated on both synthetic data and real-world data. The results show that it effectively identifies clients with noisy labels and reduces their impact on the model performance, thereby significantly outperforming existing FL approaches.
\end{abstract}

\section{Introduction}
Today, the End-Edge-Cloud ubiquitous systems, which use end sensors/devices to collect data, carry out distributed edge computing tasks, and coordinate decision support in the cloud server, have emerged to benefit many application domains \cite{ren2019survey}.
The growth of ubiquitous systems makes the collection and processing of massive amounts of personal data a possibility. This has raised privacy concerns and may hinder the development of such technologies if not addressed.

Federated Learning (FL) has emerged to be a useful machine learning paradigm to help ubiquitous systems leverage personal data in a privacy preserving manner \cite{FL:2019}.
Under FL, multiple clients collaborate to train an FL model without exchanging raw data.
It has been applied in ubiquitous systems are shown promising results \cite{wang2019edge}.
Nevertheless, one key challenge that remains open and hinders wide spread adoption of FL in ubiquitous systems, especially in the healthcare domain, is label quality disparity.

The quality of labels in clients’ local datasets influences the performance of the FL model. Existing FL approaches implicitly assume that there is no significant difference among the quality of labels from local datasets \cite{Kairouz-et-al:2019}.
Thus, popular FL approaches such as \texttt{FedAvg} treat model parameters from different clients equally \cite{McMahan-et-al:2016}.
Due to difference in the annotators’ skills, biases or malicious tampering, label noise is common in data collected by ubiquitous systems \cite{zhang2019personal,zeni2019fixing}.
Taking healthcare as an example, there are generally more cases of miss-diagnosis in smaller hospitals than in larger more well-staffed hospitals.
In FL, a noisy client can negatively impact the learned model \cite{Kairouz-et-al:2019}.
Therefore, enabling FL to effectively deal with label quality disparity is of vital importance to its success in ubiquitous systems. In this paper, we propose the Federated Opportunistic Computing for Ubiquitous System (FOCUS) approach to address this challenging problem. It is designed to identify clients with noisy labels and aggregating their model parameters into the FL model in an opportunistic manner.

FOCUS works for the cross-silo federated settings. It maintains a small set of benchmark samples in the FL server. During the FL model training process, a local model which is trained on the local data in the clients and the FL model which is the aggregated model on the FL server form a Twin Network.
By defining a contrastive loss of the Twin Network, the credibility of each client’s data can be measured.
It is then used to determine the extent to which a given client is allowed to participate in FL. 
In each iteration, FOCUS performs credibility-weighted orchestration on the FL server to avoid update corruption.
The term ``Opportunistic'' is used to indicate that a client model is not aggregated into the FL model by simple averging (as in the case of \texttt{FedAvg} \cite{mcmahan2017communication}, but weighted by its credibility.

To evaluate FOCUS, we firstly test it on a synthetic human activity recognition dataset in which labels are tampered in different ways in a subset of the clients. 
Then, it is tested on a real-world dataset from hospitals with diverse label qualities for detecting Parkinson’s Disease symptoms.
The experiment results show that FOCUS can detect clients with noisy labels and reduce their impact on the FL model performance more effectively compared to existing FL approaches.

\section{Related Work}
Label noise is common problem in machine learning, especially for deep learning on large datasets. 
There are two categories of methods to deal with this problem: 1) at the data level and 2) at the algorithm level.

At the data level, existing methods generally aim to sanitize the noisy labels to mitigate their impact.
\cite{cretu2008casting} uses small slices of the training data to generate multiple models and produce provisional labels for each input.
This is used to determine if noisy labels are present.
\cite{xie2019zeno} designed Byzantine-robust aggregators to defend against label-flipping data poisoning attacks on convolutional neural networks.
However, \cite{koh2018stronger} recently found that a federated approach to data sanitizaiton is still vulnerable to data poisoning attacks.

At the algorithm level, existing methods generally aim to train noise-tolerant models.
\cite{natarajan2013learning} studied the impact of label noise in binary classification from a theoretical perspective, and proposed a simple weighted surrogate loss to establish a strong empirical risk bounds.
Since deep learning models can easily overfit to the label noise, \cite{zhang2019metacleaner} used meta-learning to train deep models, where synthetic noisy labels were generated to update the model before the conventional gradient update.
Nevertheless, these existing methods cannot be directly applied in the context of federated learning as they require access to raw data. 

In FL, label noise is also related to non-IID issue.
\cite{zhao2018federated} found that the non-IID clients produced a poor global model in FL since the large Earth Mover's Distance (EMD) among the clients' data made their models diverse.  
However, the proposed data sharing strategy requires more communication and risks diluting the clients' information. Furthermore, the calculation of EMD requires the FL server to have access to clients raw data, which is not permissible under FL settings.

To the best of our knowledge, there is currently no published work on mitigating the impact of label noise under FL settings.

\section{Traditional FL Model Training}
Under FL, the training data are distributed among $K$ clients, each storing a subset of the training data $\mathcal{D}_k=(\mathcal{X}_k, \mathcal{Y}_k)$, $k=1,\dots,K$. Each client trains its local model $\mathcal{M}^k$ by minimizing the loss function on its own dataset only. 

Many different machine learning algorithms can be trained with FL \cite{FL:2019}. For simplicity of exposition, we use the convolutional neural networks (CNN) architecture as the basis to train an FL classification model in this paper. In this context, the cross entropy as the objective function which needs to be minimized:
\begin{equation}
\mathcal{L} = -\frac{1}{n_k}\sum_{i=1}^{n_k} y_i\log P(y_i|x_i).
\label{eqn:cross}
\end{equation}
$|n_i|$ denotes the amount of the training data owned by the $i$-th client.
After that, the FL server collects the model updates from the clients, and aggregates them to form the new global FL model $\mathcal{M}^s$. 

The most widely used FL aggregation method is the Federated Averaging \texttt{FedAvg} algorithm \cite{McMahan-et-al:2016}, which is given by:
\begin{equation}
\mathcal{M}^s_{t} = \sum_{k=1}^{K} \frac{n_k}{n} \mathcal{M}_{t}^k 
\label{eqn:fedavg}
\end{equation}
where $\mathcal{M}_{t}$ denotes the global model weight updates, $n$ is the total amount used for FL model training by the clients involved, $n=\sum_{k=1}^K n_k$.

\section{The Proposed \textnormal{FOCUS} Approach}
The proposed FOCUS approach quantifies label noise in the dataset from each FL client under horizontal federated learning. It measures the quality of each client's data and aggregates their local model updates into the FL model in an opportunistic manner. For clarity, we only present the case where each client sends local model updates to the server in plaintext. Nevertheless, added protection mechanism, such as homomorphic encryption and secret sharing, can be added into FOCUS following methods explained in \cite{FL:2019}.


\begin{figure*}[t]
    \centering
    \includegraphics[width=0.75\linewidth]{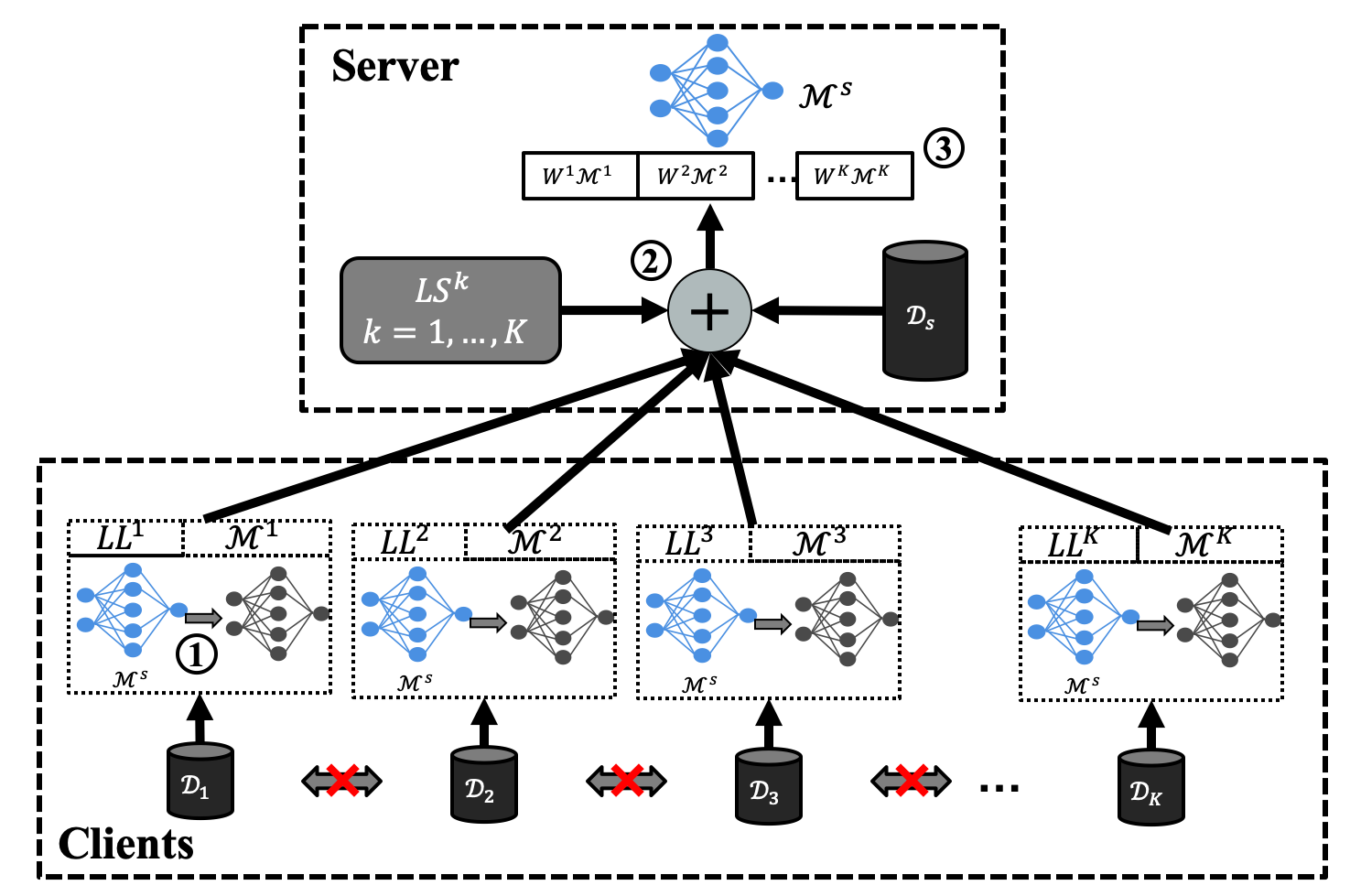}
    \caption{The pipeline of FOCUS}
    \label{FOCUS}
\end{figure*}

The pipeline of FOCUS is shown in Figure \ref{FOCUS}.
Once the $K$ clients have sent the local models to the FL server, and each client has received the global FL model from the FL server:
\begin{enumerate}
    \item Each client $i$ evaluates the global FL model on its own local dataset, and sends the evaluation result, $LL^{i}$, to the FL server.
    \item The FL server evaluates each client $i$'s local model $M^{i}$ one by one on its benchmark dataset and records the model performance as $LS^{i}$.
    \item Once the corresponding $LL^{i}$ value is received by the FL server, it computes the cross entropy between $LL^{i}$ and $LS^{i}$ to produce a credibility measure which reflects the quality of client $i$'s local labels.
    \item Finally, the credibility measure for each client $i$ is used as its weight in a weighted FedAvg operation to produce a new global FL model.
\end{enumerate}
In the following parts of this section, we provide more details on the FOCUS pipeline.

\subsection{Client Label Noise Measurement}
Since there is no prior knowledge about the clients' annotation quality, FOCUS maintains a small set of benchmark samples $\mathcal{D}_s = (\mathcal{X}_s,\mathcal{Y}_s)$ in the FL server.
A user adopting FOCUS needs to ensure that there is little noise in the benchmark dataset (i.e. data are labeled accurately). This may require the adopter to work closely with specialists in the target field, which is beyond the scope of this paper. Once this requirement is satisfied, we have the following theorem:
\begin{theorem}
\label{the:one}
Given a benchmark dataset $\mathcal{D}_s$ in the FL server, data a client $\mathcal{D}_k$ follows an identical distribution to the benchmark dataset if the trained local model $\mathcal{M}^k$ performs well on $\mathcal{D}_s$.
\end{theorem}

As $\mathcal{D}_s$ has accurate annotations, a similar data distribution indicates that the client dataset also has accurate annotations.
However, its inversion is not always correct as due to potential concept drift. In this paper, we do not address this issue.

To measure clients' label noise accurately, FOCUS considers information about how the global FL model performs on a given local dataset.
For this purpose, we define a mutual cross-entropy between the global FL model and the local model from each client to quantify the latent probability of noise, which is given by:
\begin{eqnarray}
E^k &=& LS^k + LL^ k \\
LS^k &=&-\sum_{(x,y)\in\mathcal{D}_s}y\log P(y|x;\mathcal{M}^k)\\
LL^ k &= &- \sum_{(x,y)\in\mathcal{D}_k}y\log P(y|x;\mathcal{M}^s)
\end{eqnarray}
$E^k$ combines client $k$'s local model performance on the benchmark dataset ($LS^k$) and the performance of the global FL model on client $k$'s local dataset ($LL^k$).
There are three possible cases when analyzing $E^k$:
\begin{itemize}
	\item \textbf{Small $E^k$}: A small $E^k$ indicates that the local data follows similar distribution as the benchmark dataset, meaning that client $k$'s dataset possess accurate labels.
	\item \textbf{Large $E^k$}: If both of the global FL model and the local model perform badly when tested on each other's dataset, it will result in a large $E^k$. This means either the client's dataset follows a different data distribution compared to the benchmark dataset. Thus, client $k$ is likely to possess noisy labels.
	\item \textbf{Medium $E^k$}: If either one of the two the models performs badly, it will lead to a medium $E^k$ value. In this case, it is not a sufficient to determine that client $k$ has noisy labels. If the local model is the one with poor performance, it means that the local dataset is not large enough to train a good model. If the global FL model is the one with poor performance, it means that there one or more other clients which contributed to training the FL model may have noisy labels.
	Even when a client $k$ artificially inflates the $LL^{k}$ value he sends to the FL server, the resulting $E^k$ value will be categorized into this case, and its impact on FL model performance will be limited.
\end{itemize}

Based on the mutual cross-entropy, we define a client $k$'s \textit{credibility} $C^{k}$, which reflects the quality of their local data labels, as:
\begin{equation}
C^k = 1- \frac{e^{\alpha E^k}}{\sum_{i} e^{\alpha E^i}}.
\end{equation}
$\alpha$ is a hyper-parameter for normalization.
With this measure, we propose a new algorithm to aggregate model updates from clients based on their credibility values to improve \texttt{FedAvg}.

\subsection{Opportunistic FL Model Update}
To leverage the measured client credits, we rewrite the \texttt{FedAvg} model update rule from Eq. \eqref{eqn:fedavg} as:
\begin{eqnarray}
\mathcal{M}^s_{t} = \sum_{k=1}^K W^k_{t-1} \mathcal{M}^k_{t}.
\end{eqnarray}
The term related to the amount of data involved is combined with clients' credibility values, which may vary in different rounds.
Given the client credits $C_{t}^k$ is assigned to client $k$ in round $t$, $W^k_{t}$ is defined as:
\begin{equation}
W^k_{t} = \frac{n_k C_{t}^k}{\sum_{i=1}^K n_i C_{t}^i}.
\label{eqn:w}
\end{equation}
As the mutual cross-entropy is based on both the local models and the global one at round $t$, the opportunistic updating is weighted by the latest credibility values which is calculated at rount $t-1$.
Note that since $\sum_{k=1}^K W^k_{t+1} =1$, the convergence of the proposed FOCUS approach is guaranteed as long as the \texttt{FedAvg} algorithm in Eq. \eqref{eqn:fedavg} converges.
The proposed FOCUS approach is shown as Algorithm \ref{alg:focus}.
The function \textbf{ModelTest} which is used to calculate the cross-entropy loss of a model $\mathcal{M}$ on a dataset $\mathcal{D}$ is shown as Algorithm \ref{alg:test}.

\begin{algorithm}[!t]
\caption{FOCUS}
\label{alg:focus}
\textbf{FL Server executes:}
\begin{algorithmic}[1] 
\STATE Initialize $\mathcal{M}^s_0$, $W^k_0=\frac{n_k}{\sum_k n_k}$
\FOR{each round $t=1,\dots, T$}
\FOR{each client $k=1,\dots, K$ \textbf{in parallel}}
\STATE $\mathcal{M}^k_{t} \leftarrow \textbf{ClientUpdate}(k, \mathcal{M}^s_{t})$
\STATE $LS^k_{t} \leftarrow \textbf{ModelTest}(\mathcal{M}^k_{t}, \mathcal{D}_s)$
\ENDFOR
\STATE $\mathcal{M}^s_{t}\leftarrow\sum_{k=1}^KW^k_{t-1}\mathcal{M}^k_{t}$
\FOR{each client $k=1,\dots, K$ \textbf{in parallel}}
\STATE $LL^k_{t}\leftarrow\textbf{ModelTest}(\mathcal{M}^s_{t}, \mathcal{D}_k)$
\STATE $E^k_{t}\leftarrow LS^k_{t}+LL^k_{t}$
\ENDFOR
\STATE $C^k_{t}\leftarrow1- \frac{e^{\alpha E^k}}{\sum_{i} e^{\alpha E^i}}$ 
\STATE $W^k_{t}\leftarrow  \frac{n_kC_{t}^k}{\sum_{i=1}^K n_iC_{t}^i}$
\ENDFOR
\end{algorithmic}
\textbf{ClientUpdate($k,\mathcal{M}$):}
\begin{algorithmic}[1]
\FOR{local step $j=1,\dots,N$}
\STATE $\mathcal{M}\leftarrow \mathcal{M}-\eta\nabla f(\mathcal{M};x,y)$ for $(x,y) \in \mathcal{D}_k$ 
\ENDFOR
\RETURN $\mathcal{M}$ to server
\end{algorithmic}
\end{algorithm}

\begin{algorithm}[!t]
\caption{ModelTest}
\label{alg:test}
\textbf{ModelTest}$(\mathcal{M},\mathcal{D})$:
\begin{algorithmic}[1]
\FOR{$(x,y)\in \mathcal{D}$}
\STATE $l = l-y\log p(x|y; \mathcal{M})$
\ENDFOR
\RETURN $l$
\end{algorithmic}
\end{algorithm}
\subsubsection{Communication Cost} 
FOCUS requires two communications per round: 1) broadcasting the global model, and 2) clients submit local model parameter updates to the FL server for aggregation. 
During broadcast, the central server sends $\mathcal{M}^s$ to all the clients. 
During aggregation, all or part of the $K$ clients send their local model parameters, $(LL^k,\mathcal{M}^k),k=1,\dots,K$, to the FL server.
Compared with \texttt{FedAvg}, the only item that needs to be transmitted in addition model parameters is the performance value of the global FL model on each local dataset.

\section{Experimental Evaluation}
In this section, we report experimental evaluation results showing the advantages of FOCUS over existing approaches based on real-world datasets in the cross-silo scenarios.

\subsection{Experiment Settings}
Two healthcare-related datasets are adopted in our experiments. They are:
\begin{itemize}
    \item \textbf{USC-HAD} \cite{zhang2012usc}: This dataset is a public benchmark dataset for human activity recognition, which contains 12 most common types of human activities in daily life from 14 subjects. 
    The activity data were captured by a 9-axis inertial sensor worn by the subjects on their front right hip.
    The data were collected over 84 hours.
    \item \textbf{PD-Tremor}\cite{chen2017pdassist}: This dataset was collected from Parkinson's Disease (PD) patients by measuring their tremor, which is one of the most typical motor symptoms.
    The subject was required to hold a smartphone for 15 seconds in a relaxing status.
    The hand motion data were collected by sensors embedded in the smartphone including the accelerometer, the gyroscope and the magnetic sensor. 
    Data were collected from 99 subjects in 3 hospitals.
\end{itemize}

A sliding-window is employed to segment the data stream. The window width is set to 1 second with no overlap.
A convolutional neural networl (CNN) is used to train a model based on the datasets through Stochastic Gradient Descent (SGD).
Each axis of the multi-modal sensors is regarded as a channel.

We compare FOCUS with the popular \texttt{FedAvg} approach in our experiment.
The dataset is split into the training set and the testing set.Accuracy of the global FL model is used as the evaluation metric, which is calculated as:
\begin{equation}
    Accuracy = 1 - \frac{\sum_{i=1}^N [y_i\neq\bar{y}_i]}{N}
\end{equation}
where $\bar{y}_i$ denotes the predicted label of sample $i$.

\subsection{Evaluation on the Synthetic Dataset}
In this experiment, we study the negative impact of noisy clients in federated learning.
USC-HAD is a dataset in which all the samples are correctly annotated.
To simulate a federating learning setting, the whole dataset is divided into 5 parts. One part is selected at random to be the benchmark dataset on the FL server under FOCUS. The others are distributed to the clients.
Then, one of the 4 clients is randomly selected to be the noisy client. The labels in this client are randomized.

\begin{figure}[!b]
    \centering
    \includegraphics[width=1\columnwidth]{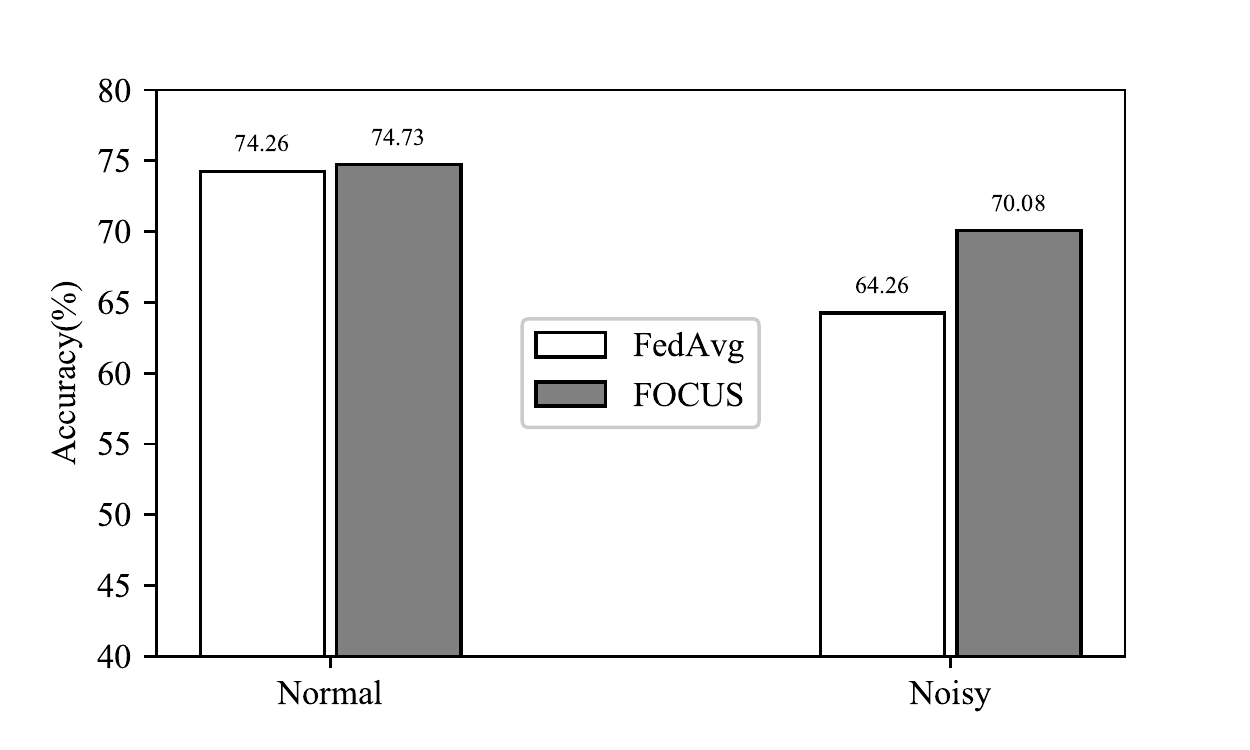}
    \caption{The test accuracy comparison on USC-HAD}
    \label{fig:result}
\end{figure}

There are two scenarios for federating learning. One of which is referred to as ``Normal'', where all the four clients are annotated with correct labels. The other is referred to as ``Noisy'', where one of clients has noisy labels.
The testing accuracy comparison between \texttt{FedAvg} and FOCUS under these two scenarios is shown in Figure \ref{fig:result}.

\begin{figure}[!b]
    \centering
    \includegraphics[width=1\columnwidth]{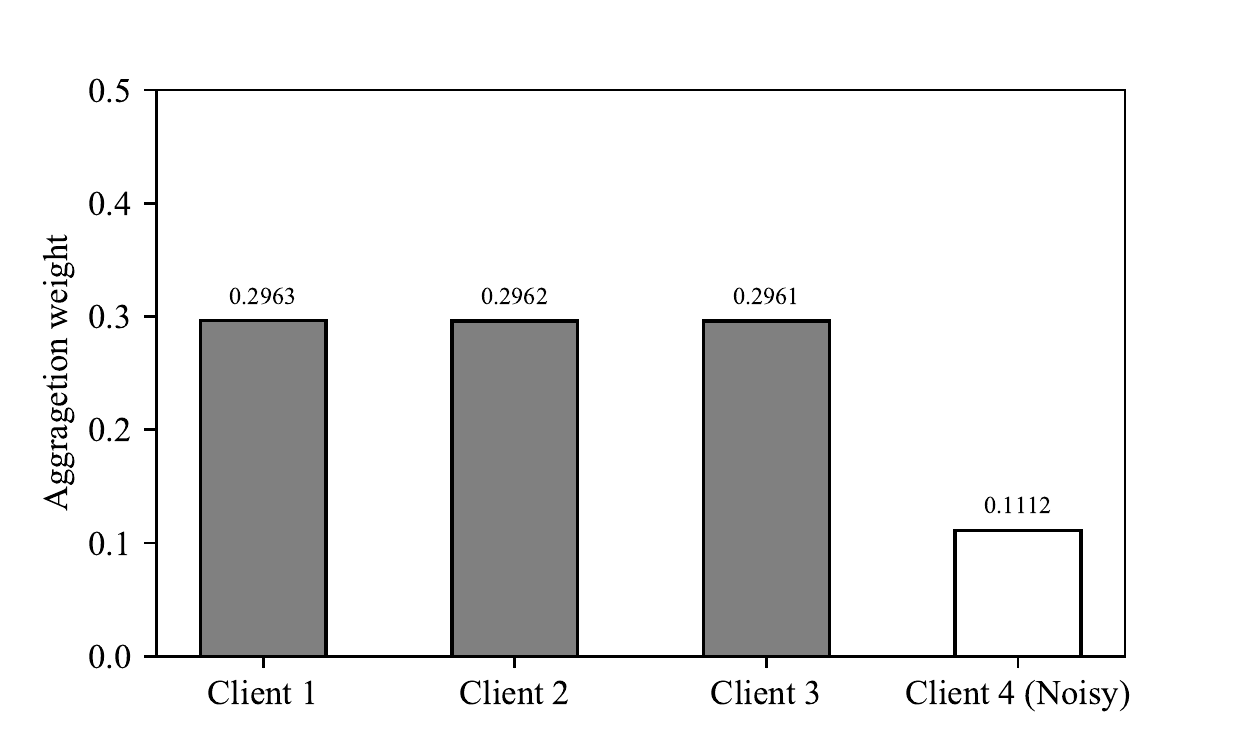}
    \caption{The weights assigned to the clients by FOCUS}
    \label{fig:weights}
\end{figure}

It can be observed that under the Normal scenario, FOCUS and FedAvg achieved the same performance in terms of accuracy. In this sense, \texttt{FedAvg} can be regarded as a special case of FOCUS, which does not take the credibility of clients into account. 

Under the Noisy scenario, due to the noisy client, some valuable information is lost and performance degradation is significant for both \texttt{FedAvg} and FOCUS.
Since all the local models including that from the noisy client are aggregated indiscriminately in \texttt{FedAvg}, its performance is significantly poorer than FOCUS.
Through noisy client detection and opportunistic model aggregation, FOCUS outperforms \texttt{FedAvg} by 5.82\% in terms of accuracy.


The opportunistic aggregation weights produced by FOCUS for the 4 clients in the last learning iteration are shown in Figure \ref{fig:weights}.
As the data on normal clients follows an identical data distribution, they are assigned almost equal weight during FL model aggregation.
However, the weight for the noisy client which has been significantly reduced by FOCUS, which shows that the proposed method can correctly detect noisy client and take appropriate actions.

\begin{figure}[!t]
    \centering
    \includegraphics[width=1\linewidth]{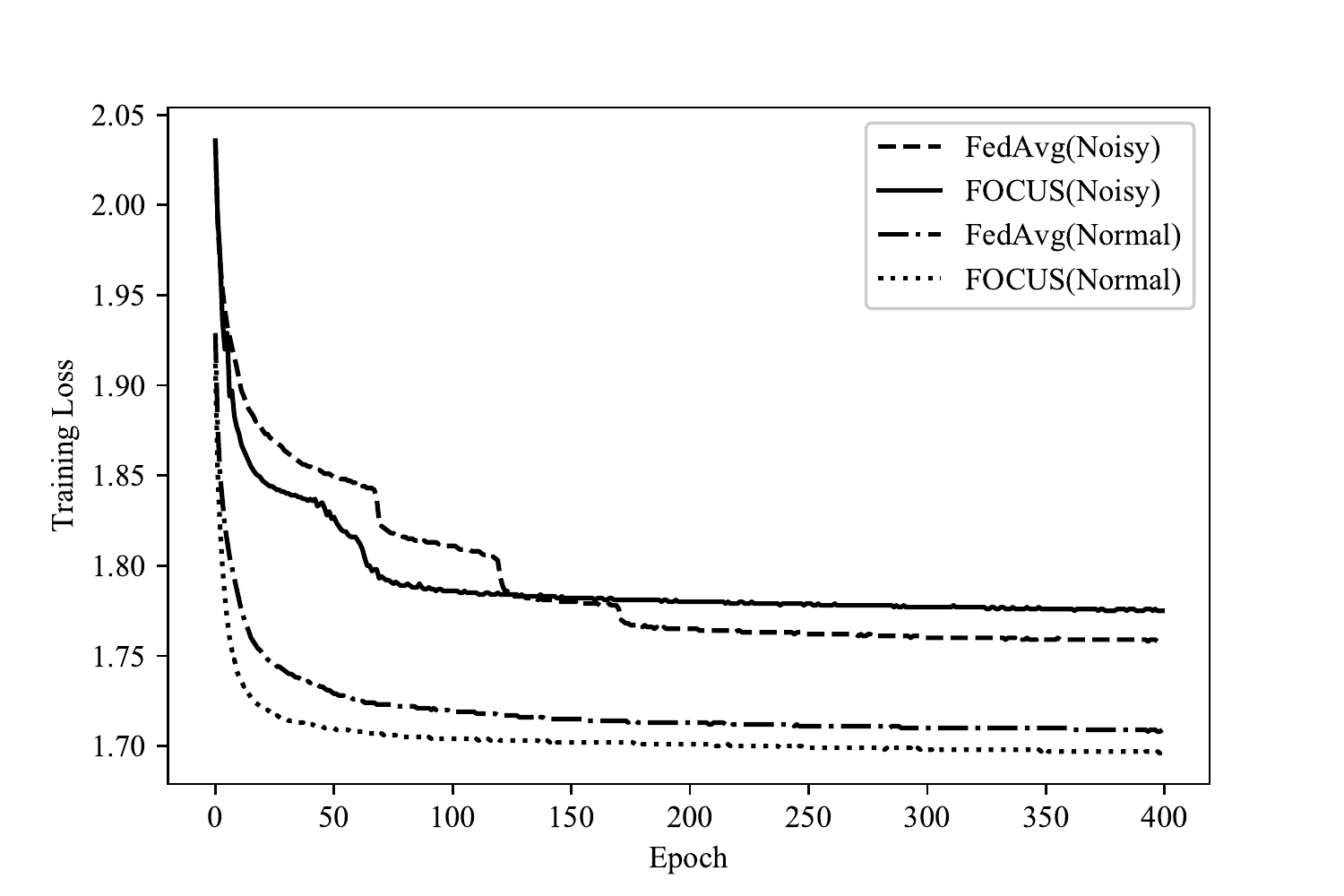}
    \caption{The training loss comparison results on USC-HAD}
    \label{fig:loss}
\end{figure}

Figure \ref{fig:loss} shows the training loss comparison during each FL training iteration with a same learning rate setting.
The training loss at round $t$ is calculated as:
\begin{equation}
    l^t_{fl}= \frac{1}{K}\sum_{k=1}^K \mathcal{L}(\mathcal{M}_t^k,\mathcal{D}_k)
\end{equation}
$\mathcal{L}$ is the cross-entropy loss in our experiments.

Both \texttt{FedAvg} and FOCUS take longer to converge under the Noisy scenario compared to under the Normal scenario.
Nevertheless, the convergence rate of FOCUS under both scenarios is faster than that of \texttt{FedAvg}.
Because of the incorrect labels, the data distribution of the noisy clients are different from the others, resulting in larger Earth Mover's Distance values and diverse model parameters \cite{zhao2018federated}.
Thus, during the aggregation in the server under FOCUS, the global model is less impacted by the noisy client due to its reduced weight.

Another evidence for the reduced impact of the noisy client on the FL model is that the final loss achieved by FOCUS is larger than that of \texttt{FedAvg}.
This is because the the global FL model under FOCUS does not fit the noisy data as well as the normal data. This results in a larger training loss on the noisy data.
In other words, FOCUS is capable of avoiding over-fitting the noisy data.

\subsection{Evaluation on the Real-world Dataset}
In this section, we evaluate FOCUS onto a real-world practical dataset for Parkinson's Disease symptom recognition - PD-Tremor - by comparing it with the popular \texttt{FedAvg} approach.

Among the 3 hospitals from which this dataset was collected, 2 of them are top-tier hospitals, and the third one is considered a lower-tier hospital.
We regard each hospital as a client in FL.
All the data are annotated by doctors from the 3 hospitals.
As doctors from the lower-tier hospital tend to be less experienced and are more likely to make wrong annotations, we test FOCUS on this dataset to evaluate its effectiveness.

To collect a set of benchmark samples, two experts were invited to make consistent annotations on a sample dataset.
The benchmark samples are divided into two parts. One of them is used as the benchmark dataset on the FL server; and the other is used as the test set.

\begin{figure}[ht]
    \centering
    \includegraphics[width=1\linewidth]{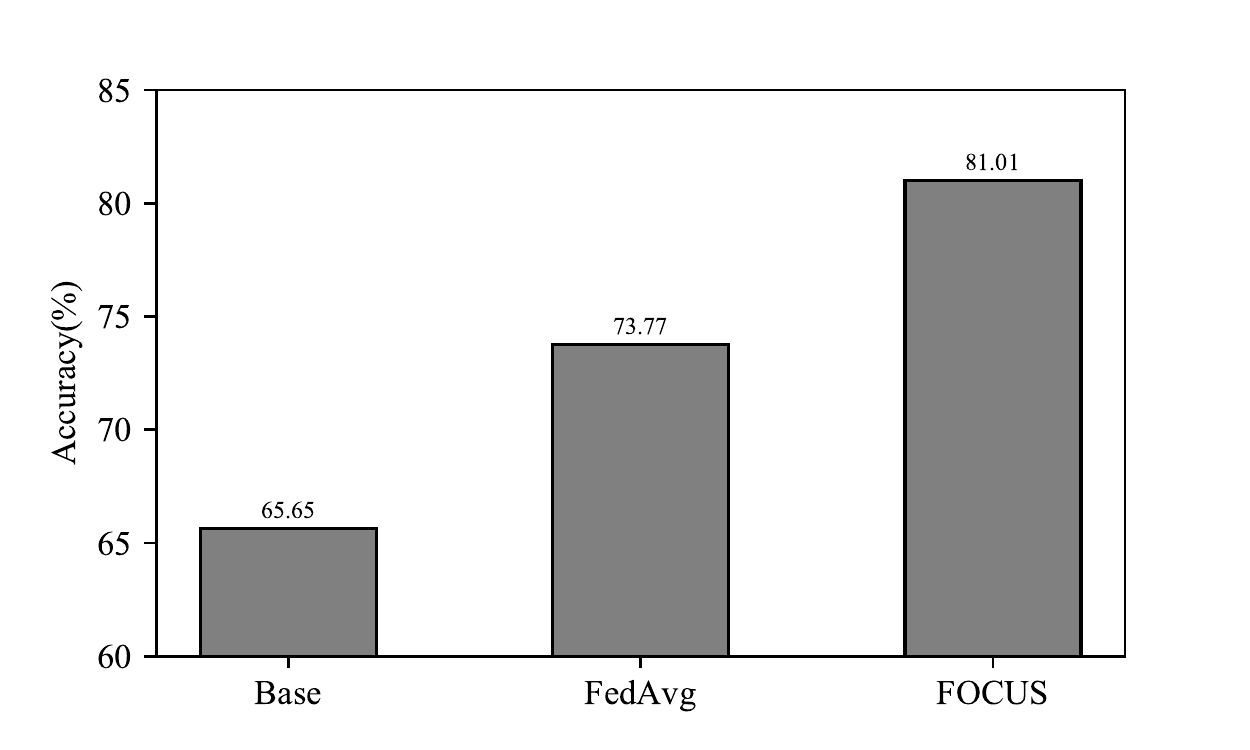}
    \caption{The test accuracy comparison on PD-Tremor}
    \label{fig:resultpd}
\end{figure}

\begin{figure}[ht]
    \centering
    \includegraphics[width=1\linewidth]{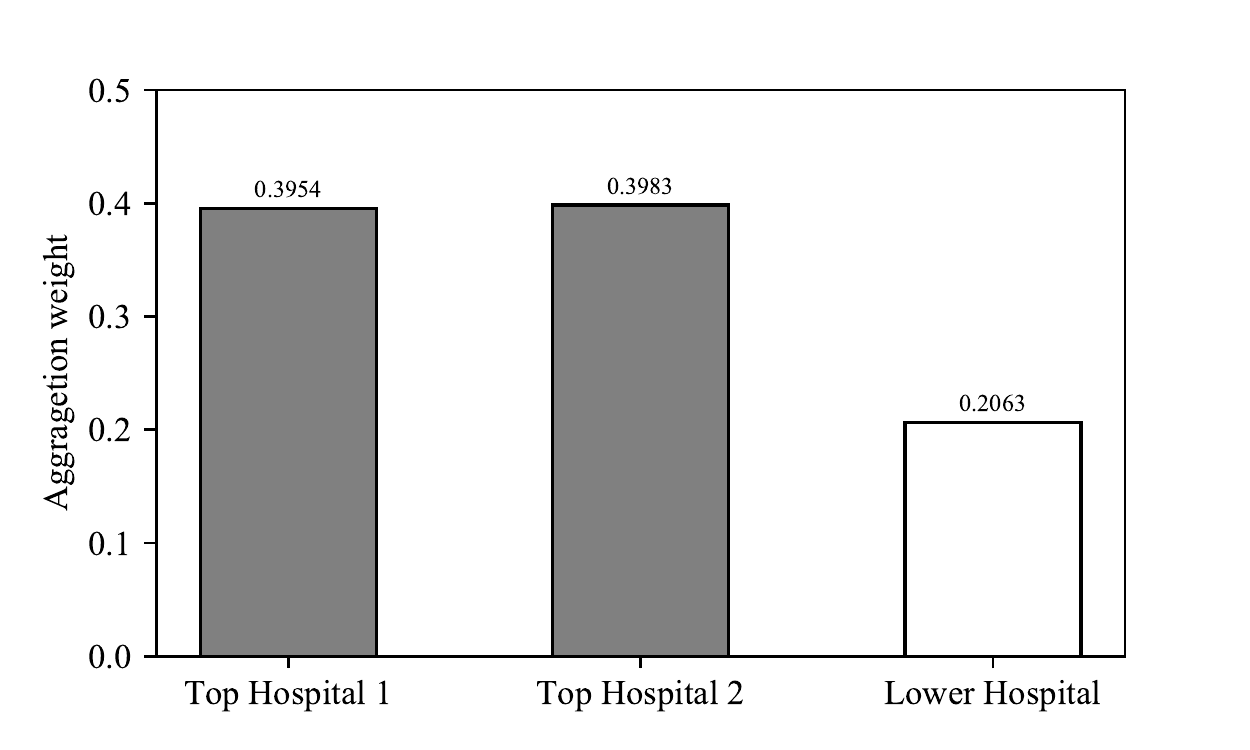}
    \caption{The weights assigned to the hospitals by FOCUS}
    \label{fig:weightpd}
\end{figure}

The results in terms of prediction accuracy and client weights in FOCUS are shown in Figures \ref{fig:resultpd} and \ref{fig:weightpd}, respectively.
``Base'' denotes the base model trained with the benchmark dataset.

It can be observed that both FL-based approaches, FOCUS and \texttt{FedAvg}, are able to learn more information from the clients and train stronger models. FOCUS outperformed \texttt{FedAvg} in terms of accuracy by 7.24\%, which also confirmed our suspicion that there are noisy are in the clients.
By observing the opportunistic weight of each client, we find that the lower-tier hospital is assigned a smaller weight, which indicates that its data are of low-quality and contain noisy labels.


In summary, FOCUS has significantly outperformed the state-of-the-art \texttt{FedAvg} algorithm under both the synthetic dataset and the real-world dataset.

\section{Conclusions and Future Work}
Label quality disparity is an important challenge facing today's federated learning field. So far, it remains open. Noisy labels in FL clients can corrupt the learned FL model. Since under FL, sensitive local data cannot be transmitted out of the owner client's data store in order to protect user privacy. This makes the problem of noisy local labels even more challenging to resolve. In this paper, we propose the Federated Opportunistic Computing for Ubiquitous System (FOCUS) to address this challenging problem.
FOCUS maintains a small set of benchmark samples in the server.
A novel mutual cross-entropy based credibility score is designed to compute the label quality of a client's dataset without requiring access to raw data.
Based on the measured credibility, we further proposed a modification to the popular \texttt{FedAvg} algorithm to opportunistically aggregate client model updates into a global FL model.
In this way, only a parameter which carries the local loss is extra communicated.
Extensive experiments on both synthetic and real-world data demonstrated significant advantage of FOCUS over \texttt{FedAvg}. With FOCUS, we empower FL systems to effectively identify clients with noisy label and improve their model training strategy to mitigate the negative effects. To the best of our knowledge, it is the first FL approach capable of handling label noisy in a privacy preserving manner.

Although FOCUS is proved to be effective in the federated learning with label quality disparity, there are still interesting problems which require further investigation.
For example, how to distinguish clients who maliciously attack the FL system by faking their labels and the those facing genuine difficulties in providing correct labels is an important issue which affects how these clients should be dealt with. In subsequent research, we will focus on tackling this problem.

\section*{Acknowledgments}
Thank the clinical doctors from the China Parkinson Alliance for supporting the data acquisition.
This work was supported by National Key Research \& Development Plan of China No.2017YFB1002802; Natural Science Foundation of China No.61972383; Research \& Development Plan in Key Field of Guangdong Province No.2019B010109001.

\bibliographystyle{named}
\bibliography{ijcai20}

\end{document}